\documentclass[lettersize,journal]{IEEEtran}
\usepackage[utf8]{inputenc}
\usepackage[english]{babel}
\usepackage{multirow}
\usepackage[dvipsnames]{xcolor}
\usepackage{amsmath,amsfonts,amssymb,amscd,bm,textcomp,gensymb }
\usepackage{tikz}
\usetikzlibrary{shapes,shapes.geometric,arrows,positioning}
\usepackage{tuda-pgfplots}
\usepackage{pgfplots}
\usepgfplotslibrary{groupplots,dateplot}
\usepgfplotslibrary{groupplots}
\usetikzlibrary{patterns,shapes.arrows}
\pgfplotsset{compat=newest}
\usepackage{tabularx}
\usepackage{array}
\usepackage[caption=false,font=scriptsize,labelfont=sf,textfont=sf]{subfig}
\usepackage{textcomp}
\usepackage{stfloats}
\usepackage{url}
\usepackage{verbatim}
\usepackage{graphicx}
\usepackage{cite}
\usepackage{hyperref}
\usepackage{eso-pic}
\AddToShipoutPicture*{\tiny\sffamily\raisebox{1cm}{\hspace{1.65cm}\fbox{\parbox{\textwidth}{\copyright 2022 IEEE.  Personal use of this material is permitted.  Permission from IEEE must be obtained for all other uses, in any current or future media, including reprinting/republishing this material for advertising or promotional purposes, creating new collective works, for resale or redistribution to servers or lists, or reuse of any copyrighted component of this work in other works.}}}}
\begin{document}
\title{Performance Analysis of Electrical Machines Using a Hybrid Data- and Physics-Driven Model}

\author{Vivek Parekh, Dominik Flore, Sebastian Schöps
\thanks{Vivek Parekh: Computational Electromagnetics Group, TU Darmstadt, 64289 Darmstadt and Robert Bosch GmbH, 70442 Stuttgart, Germany (email: Vivek.Parekh@de.bosch.com).
Dominik Flore: Robert Bosch GmbH, 70442 Stuttgart, Germany (email: Dominik.Flore@de.bosch.com).
Sebastian Schöps: Computational Electromagnetics Group, TU Darmstadt, 64289 Darmstadt, Germany (email:sebastian.schoeps@tu-darmstadt.de).}}

\markboth{IEEE TRANSACTIONS ON ENERGY CONVERSION,~Vol.~xx, No.~x, xx~xxxx}%
{Shell \MakeLowercase{\textit{et al.}}: A Sample Article Using IEEEtran.cls for IEEE Journals}


\maketitle

\begin{abstract}
In the design phase of an electrical machine, finite element (FE) simulation are commonly used to numerically optimize the performance. The output of the magneto-static FE simulation characterizes the electromagnetic behavior of the electrical machine. It usually includes intermediate measures such as nonlinear iron losses, electromagnetic torque, and flux values at each operating point to compute the key performance indicators (KPIs). We present a data-driven deep learning approach that replaces the computationally heavy FE calculations by a deep neural network (DNN). The DNN is trained by a large volume of stored FE data in a supervised manner. During the learning process, the network response (intermediate measures) is fed as input to a physics-based post-processing to estimate characteristic maps and KPIs. Results indicate that the predictions of intermediate measures and the subsequent computations of KPIs are close to the ground truth for new machine designs. We show that this hybrid approach yields flexibility in the simulation process. Finally, the proposed hybrid approach is quantitatively compared to existing deep neural network-based direct prediction approach of KPIs.
\end{abstract}


\section{Introduction}

\begin{figure}
	\centering
	\input{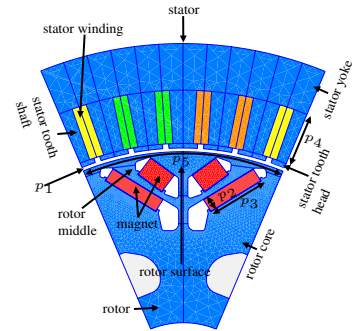}
	\vspace{-0.5em}
	\caption{Exemplary double-V PMSM geometry.}
	\label{fig:data_vis}
	\vspace{-1em}
\end{figure}

\IEEEPARstart{E}{lectrical} machines have numerous real-world applications, ranging from home appliances to the automotive industry. Permanent magnet synchronous machines (PMSM) gained popularity in recent years due to their high efficiency, large power density, and high torque-current ratio, see for example \autoref{fig:data_vis}. However, the use of expensive materials such as rare earth magnets (neodymium, dysprosium, or terbium) is a significant cost operator in the manufacturing process. Therefore simulation based optimization, e.g. minimizing cost and maximizing performance, is an established technique in academia and industry \cite{Di-Barba_2010aa}. The finite element method (FEM) is at the heart of such numerical optimization workflows. In the automotive industry, FEM-based optimization has been used for decades to reduce the number of physical prototypes and to optimize the active parts of the electrical machine (stator, rotor, magnets, winding). However, it is a computationally expensive process, so there is a great need for complexity reduction.

To this end, surrogate or meta modelling has become a research field on its own in the last decade \cite{Koziel_2013aa}. More recently, meta-modelling by machine learning has gained attention. Nowadays, deep neural networks (DNNs) are employed at various design stages because of their advantages such as scalability for high dimensional data, handling of big data, easy parallelization by graphical processing units (GPUs) and automatic feature extraction. For example, Yan et al.~\cite{8921886} describes the deep-learning-based approximation of the electromagnetic torque for PMSM. In another article, the DNN is used as a meta-model for predicting objectives obtained by FE output in the design phase for the optimization of flux switching machine to lower the computational burden \cite{8785344}. Another work \cite{8961095} demonstrates how a deep learning-assisted approach is used to predict efficiency maps of electric drives. The DNN-based two-step optimization method is investigated in \cite{9366983} to speed up the design process by inferring torque-related performance. The application of CNN to approximate performance characteristics of electric machines for multidimensional outputs is investigated in \cite{act10020028}. This paper builds on the findings in \cite{9333549}. It demonstrates how a deep learning based meta-models predict cross-domain key performance indicators (KPIs) effectively using different PMSM representations in a large input design space. The meta-model trained with parameter-based data has higher prediction accuracy than an image-based representation. The generation of (high-resolution) images and the training of the deep convolutional neural network requires more time and effort than the parameter-based approach. However, the image-based meta-model remains invariant to re-parametrization that does not hold for the parameter-based model. Recently, \cite{parekh2022variational} demonstrated how KPIs of PMSM topologies with different parameterizations can be predicted concurrently by mapping high dimensional scalar parameters into a lower-dimensional regularized latent space using a variational autoencoder.

In this paper, we present a novel physics and data-driven approach for characterizing the electromagnetic behavior of an electrical machine. A multi-branch DNN functions as a meta-model. It accepts high dimensional varying scalar parameter input, e.g. geometry, electrical, and material. The results of the FE simulations are stored in a database and the DNN is trained by supervised learning. In contrast to previous approaches, we do not predict KPIs by the DNN but only intermediate measures. The result, along with the system parameters, is fed into a physics-based analytical model to calculate characteristic maps and cross-domain KPIs, e.g., maximum torque, power, cost, etc. Finally, we quantitatively compare this hybrid model to existing approaches. 

The article is organized as follows: \autoref{sec:DG} explains the problem formulation and data set specifics. \autoref{sec:MNAT} presents the methodology, network architecture, and training settings. \autoref{sec:res} provides quantitative analysis of results which is followed by the conclusion in \autoref{sec:conclusion}.

\begin{table}
	\caption{Inputs (selection)}
	\label{tab:input}
	\begin{tabularx}{\linewidth}{|c|X|r|r|c|}
		\hline
		         & \textbf{Parameter}               & \textbf{Min} & \textbf{Max}     & \textbf{Unit} \\ \hline
		$p_{1}$  & Air gap                          &   0.50       &    2.00          & mm            \\ \hline
		$p_{2}$  & Height of inner magnet           &   2.23       &    7.00          & mm            \\ \hline
		$p_{3}$  & Width of inner magnet            &   8.00       &   25.00          & mm            \\ \hline
		$p_{4}$  & Height of tooth head             &  12.00       &   20.00          & mm            \\ \hline
		$p_{5}$  & Rotor outer diameter             & 159.00       &  165.00          & mm            \\ \hline
		$p_{6}$  & Input phase current $I$          &   0.00       & 1336.40          & A             \\ \hline
		$p_{7}$  & Control angle $\alpha$           & 0            & 360              & degree        \\ \hline
		\hline
		$p_{8}$  & Input phase voltage              & \multicolumn{2}{c|}{640}        & V         \\ \hline
		$p_{9}$  & Slots per pole per phase         & \multicolumn{2}{c|}{2}          & -         \\ \hline
		$p_{10}$ & Pole pairs                       & \multicolumn{2}{c|}{4}          & -         \\ \hline
		$p_{11}$ & Stator type                      & \multicolumn{2}{c|}{symmetric}  & -         \\ \hline
		$p_{12}$ & Rotor type                       & \multicolumn{2}{c|}{double-V}   & -         \\ \hline
	\end{tabularx}%
\end{table}

\section{Problem formulation and data set}\label{sec:DG} %
Each FE simulation requires a high-dimensional parameter-based input $\mathbf{p}$ to compute the intermediate measures $\mathbf{y}$ which then leads to the KPIs $\mathbf{z}$. Let us consider an electric machine simulation dataset
\begin{equation}
    \begin{aligned}
        \mathcal{D} := \Big\{ (\mathbf{p}^{(1)},\mathbf{y}^{(1)},\mathbf{z}^{(1)}),\ldots,(\mathbf{p}^{(N)},\mathbf{y}^{(N)},\mathbf{z}^{(N)}) \Big\}
    \end{aligned}
\end{equation}
of $N_{\mathrm{D}}$ samples, where each input is generated by the realization of a $d$-dimensional random input vector $\mathbf{p}^{(i)}$ and leads to the $m$-dimensional vector $\mathbf{y}^{(i)}=\mathbf{F}(\mathbf{p}^{(i)})$ using the FE model $\mathbf{F}$ and finally to the KPIs $\mathbf{z}^{(i)}=\mathbf{K}(\mathbf{p}^{(i)},\mathbf{y}^{(i)})$ using an analytical post-processing, summarized in $\mathbf{K}$. The goal is now to learn a function $\hat{\mathbf{F}}$ that is cheap to evaluate and approximates the computationally heavy FE simulation. It will be used afterwards as a meta-model to predict the KPIs for new unseen machine designs.

\subsection{Dataset specifics}\label{sec:dataset}
In this study, the simulation dataset is generated for a double-V PMSM. The procedure for calculating KPIs is detailed in Section~2 of \cite{9333549}. A total of $N_{\mathrm{EM}}=44877$ machine designs have been simulated for 35 input parameters $p_i$, e.g. geometry and material. \autoref{fig:data_vis} depicts a representative geometry from the dataset. Each machine design was simulated using FE simulations for $N_{\mathrm{OP}}=37$ operating points, considering the magnetic state symmetry to reduce computational overhead. The operating points are interpreted as a variable electrical excitation input for the electrical machine, i.e., it is given by an input phase current $I$ and its control angle $\alpha$. 
As a result, $N_{\mathcal{D}} = N_{\mathrm{OP}} \times N_{\mathrm{EM}} = 1660449$ samples are available in the dataset on which the meta-model will be trained and tested on. A few selected input parameters are listed in \autoref{tab:input}.

\begin{figure}
\centering
    \input{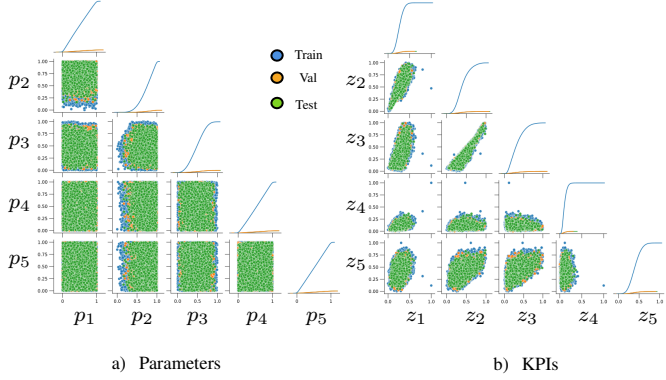}
	\vspace{-0.25em}
	\caption{Visualization of parameter and KPIs distributions.
	}
	\label{fig:DS}
\end{figure}

The simulated quantities, i.e., intermediate measures $y_i$, from the FE analysis are non-linear iron losses $P_{\mathrm{fe}}$, torque $T$ and fluxes $\Psi_{1}, \Psi_{2}, \Psi_{3}$ linked with the three coils for one electrical period, see \autoref{tab:intermediate}. Note, that the losses $P_\mathrm{fe}$ include two contributions from hysteresis and eddy currents, \cite{6689759,jordan1924ferromagnetischen}. A plot of two operating point calculations for one electrical period of a test sample from the dataset $\mathcal{D}$ is shown in \autoref{fig:evalopptsd}. Finally, examples of a few KPIs are given in \autoref{tab:kpis} and their distribution can be seen in \autoref{fig:DS}. 

\begin{table}
	\caption{Outputs (intermediate measures)}
	\label{tab:intermediate}
	\begin{tabularx}{\linewidth}{|c|X|r|r|c|}
		\hline
		         & \textbf{Measure}                         & \textbf{Unit} \\ \hline
		$y_{1}$  & Nonlinear iron losses $P_{\mathrm{fe}}$  & J             \\ \hline
		$y_{2}$  & Electromagnetic torque $T$               & Nm            \\ \hline
		$y_{3}$  & Flux linkage $\Psi_{1}$ coil 1           & Wb            \\ \hline
		$y_{4}$  & Flux linkage $\Psi_{2}$ coil 2           & Wb            \\ \hline
		$y_{5}$  & Flux linkage $\Psi_{3}$ coil 3           & Wb            \\ \hline
	\end{tabularx}%
\end{table}

\begin{table}
	\caption{Key performance indicators}
	\label{tab:kpis}
	\begin{tabularx}{\linewidth}{|c|X|l|}
		\hline
		        & \textbf{KPIs}                         & \textbf{Unit} \\ \hline
		$z_{1}$ & Maximum torque on limit curve        & Nm            \\ \hline
		$z_{2}$ & Max. shaft power                     & W             \\ \hline
		$z_{3}$ & Max. shaft power at max. speed       & W             \\ \hline
		$z_{4}$ & Max. torque ripple on limit curve    & Nm            \\ \hline
		$z_{5}$ & Material cost                        & Euro          \\ \hline
		$z_{6}$ & Mass of active parts                 & Kg            \\ \hline
		$z_{7}$ & Torque ripple deviation              & Nm            \\ \hline
	\end{tabularx}%
\end{table}

\begin{figure*}[b]
    \centering
        \subfloat[Operating point: zero current $I$ and $\alpha = 0\degree$]{\input{fig2/OP_PT_0A_0Deg}%
        \label{fig:op1d}}
        \hfil
        \subfloat[Operating point: maximal current $I$ and $\alpha = 0\degree$]{\input{fig2/OP_PT_1335.8654A_0Deg}%
        \label{fig:op2d}}
    \caption{Operating points for the test design.}
    \label{fig:evalopptsd}
\end{figure*}

\section{Procedure, Network architecture and training}\label{sec:MNAT}

This section discusses the procedure for calculating KPIs using a multi-branch DNN, followed by specifics on its architecture and training.

\subsection{Procedure}\label{subsec:MTH}

\begin{figure}
	\centering
	\scalebox{0.85}{
		\tikzstyle{block_input_parameters} = [rectangle, draw, fill=black!10, 
	text width=6.5em, text centered, rounded corners,
	minimum height=1em]
\tikzstyle{block_output_parameters} = [rectangle, draw, fill = black!10, 
	text width=6em, text centered, rounded corners,
	minimum height=1em]
\tikzstyle{block_output_kpis} = [rectangle, draw,     fill = black!10, 
	text width=5em, text centered, rounded corners,
	minimum height=2.5em]
\tikzstyle{block_small} = [rectangle, draw, fill=black!10, 
	text width=3.5em, text centered, rounded corners,
	minimum height=1em]
\tikzstyle{block_small_1} = [rectangle, draw, fill=black!10, 
	text width=5em, text centered, rounded corners,
	minimum height=2.9em]
\tikzstyle{line} = [draw, -latex']

\begin{tikzpicture}[node distance=1cm, >=stealth]
	\tikzset{font=\scriptsize}
	\begin{scope}[node distance= 2.7cm]
	\node [block_input_parameters] (init) {Varying scalar parameters (geometry, electrical, material)};
	\node [block_small_1, right of=init, label=above:{Classical approach}] (classical) {Finite element simulation};

	\end{scope}
	\begin{scope}[node distance= 2.75cm]
		\node [block_output_parameters, right of=classical] (out) {KPIs and performance curves calculation via physics- based model};
	\end{scope}
	\node [block_small, below of=init] (init_1) {System parameters};

	\path [line] (init) -- (classical);
	\path [line] (init_1.east) -|   (out.south);

	\begin{scope}[node distance= 2.3cm]
	\node [block_input_parameters, below of=init] (init_2) {Varying scalar parameters (geometry, electrical, material)};
	\end{scope}
	\begin{scope}[node distance= 2.7cm]
	    \node [block_small_1, right of=init_2,label=above:{Hybrid approach}] (MDNN) {Multi-branch DNN};
	\end{scope}
	
    \begin{scope}[node distance= 2.75cm]
	    \node [block_output_parameters, right of=MDNN] (out_2) {KPIs and performance curves calculation via physics- based model};
	\end{scope}

	\node [block_small, below of=init_2] (init_3) {System parameters};

\begin{scope}[->,>=latex]
     \draw[->] ([yshift=1 * 0.35 cm]classical.east) -- ([yshift=0.35 cm]out.west) node[left, xshift=-0.1cm,yshift=0.13cm]  (oc) [font=\fontsize{0.41em}{0.56em}\selectfont]{$P_{\mathrm{fe}}$};
     \draw[->] ([yshift=1 * 0 cm]classical.east) -- ([yshift=0 cm]out.west) node[left, xshift=-0.1cm,yshift=0.13cm]  (od)[font=\fontsize{0.41em}{0.56em}\selectfont] {$T$};
     \draw[->] ([yshift=-1 * 0.35 cm]classical.east) -- ([yshift=-0.35 cm]out.west) node[left, xshift=-0.1cm,yshift=0.14cm] (o4)[font=\fontsize{0.41em}{0.56em}\selectfont]{$\Psi$} ;

     \node(d0) at ([xshift=0.5cm, yshift=0.4cm]out.east) {$z_1$};
     \node(d1) at ([xshift=0.5cm, yshift=0.2 cm]out.east) {$z_2$};
     \node(d2) at ([xshift=0.5cm, yshift=0.0 cm]out.east) {$\vdots$};
     \node(d3) at ([xshift=0.5cm, yshift=-0.4 cm]out.east) {$z_n$};
     \draw [->] ([yshift=0.4 cm]out.east) -- (d0);
     \draw [->] ([yshift=0.2 cm]out.east) -- (d1);
     \draw [->] ([yshift=-0.4 cm]out.east) -- (d3);

     \node(v0) at ([xshift=1cm, yshift=-0.2cm]init.south){};
     \draw [-] ([xshift=1cm, yshift=0.0cm]init.south) -- (v0);
     \draw [->] (v0) |- ([yshift=-0.6 cm]out.west);
     
\end{scope}

\begin{scope}[->,>=latex]
  \draw[->] ([yshift=1 * 0.35 cm]MDNN.east) -- ([yshift=0.35 cm]out_2.west) node[left, xshift=-0.1cm,yshift=0.13cm] (o1) [font=\fontsize{0.41em}{0.56em}\selectfont]{$\hat{P}_{\mathrm{fe}}$};
  \draw[->] ([yshift=1 * 0 cm]MDNN.east) -- ([yshift=0 cm]out_2.west)     node[left, xshift=-0.1cm,yshift=0.13cm] (o2)[font=\fontsize{0.41em}{0.56em}\selectfont] {$\hat{T}$};
  \draw[->] ([yshift=-1 * 0.35 cm]MDNN.east) -- ([yshift=-0.35 cm]out_2.west) node[left, xshift=-0.1cm,yshift=0.14cm] (o4)[font=\fontsize{0.41em}{0.56em}\selectfont]{$\hat{\Psi}$} ;
  
 \node(p0) at ([xshift=0.5cm,yshift=0.4 cm]out_2.east) {$\hat{z}_{1}$};
 \node(p1) at ([xshift=0.5cm,yshift=0.2 cm]out_2.east) {$\hat{z}_{2}$};
 \node(p2) at ([xshift=0.5cm,yshift=0.0 cm]out_2.east) {$\vdots$};
 \node(p3) at ([xshift=0.5cm,yshift=-0.4 cm]out_2.east) {$\hat{z}_{n}$};
    \draw [->] ([yshift=0.4 cm]out_2.east) -- (p0);
    \draw [->] ([yshift=0.2 cm]out_2.east) -- (p1);
    \draw [->] ([yshift=-0.4 cm]out_2.east) --(p3);
    
\end{scope}

	\path [line] (init_2) -- (MDNN);

	\path [line] (init_3.east) -|   (out_2.south);
	\node(s0) at ([xshift=1cm, yshift=-0.2cm]init_2.south){};
     \draw [-] ([xshift=1cm, yshift=0.0cm]init_2.south) -- (s0);
     \draw [->] (s0) |- ([yshift=-0.6 cm]out_2.west);
	\begin{scope}[node distance= 2.3cm]
	\node [block_input_parameters, below of=init_2] (init_4) {Varying scalar parameters (geometry, electrical, material)};
	\end{scope}
	\begin{scope}[node distance= 2.75cm]
	\node [block_output_kpis, right of=init_4, label=above:{Direct DL approach \cite{9333549}}] (DNN) {KPIs prediction via DNN};
	\end{scope}
    \begin{scope}[->,>=latex]
      \node(k0) at ([xshift=0.5cm,yshift=0.28 cm]DNN.east) {$\hat{z}_{1}$};
      \node(k1) at ([xshift=0.5cm,yshift=0.08 cm]DNN.east) {$\hat{z}_{2}$};
      \node(k2) at ([xshift=0.5cm,yshift=-0.08 cm]DNN.east) {$\cdot$};
      \node(k3) at ([xshift=0.5cm,yshift=-0.28 cm]DNN.east) {$\hat{z}_{n}$};
      \draw [->] ([yshift=0.28 cm]DNN.east)   -- (k0);
      \draw [->] ([yshift=0.08 cm]DNN.east)  -- (k1);
      \draw [->] ([yshift=-0.28 cm]DNN.east) -- (k3);
    \end{scope}
	\path [line] (init_4) -- (DNN);
	
\end{tikzpicture}
\label{fig:FCP}
	}
	\caption{Block diagrams of different approaches for calculation of KPIs.
	}
	\label{fig:MTH}
\end{figure}
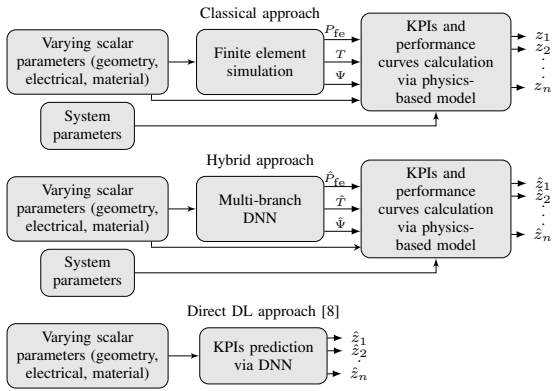

\autoref{fig:MTH} shows block diagrams of three different approaches for the calculation of KPIs. The classical approach is a conventional way of computing KPIs and is widely used in industry. In this method, as shown in the diagram, the input parameters $p_i$ are fed to the FE simulation, and then the result $y_i$, along with the inputs $p_i$, are given to a physics-based in-house post-processing tool to estimate final KPIs $z_i$. The true functions for computing FE outcomes and KPIs can be written abstractly as,
\begin{equation}
	\mathbf{z}=\mathbf{K}\bigl(\mathbf{p},{\mathbf{F}}(\mathbf{p})\bigr).
	\label{eq:true_KF}
\end{equation}
Until now classical data-driven approaches, e.g. \cite{9333549}, proposed to learn the function composition  $\textbf{p}\to\mathbf{z}$, i.e.
\begin{equation}
	\mathbf{z}\approx\hat{\mathbf{K}}_\varphi\bigl(\mathbf{p}\bigr)
	\label{eq:approx_K}
\end{equation}
where $\varphi$ represents DNN model parameters. Here, the approximation $\hat{\mathbf{K}}_{\varphi}$ works as a meta model to directly predict the KPIs. This model is trained by optimizing the model parameters $\varphi$, i.e., minimizing the difference between the ground truth and the prediction. The training loss function using the $\ell_{1}$-norm can be written as
\begin{equation}
	\min_\varphi {L}(\varphi) :=
	\sum_i \|\mathbf{z}^{(i)} - \hat{\mathbf{K}}_{\varphi}(\mathbf{p}^{(i)}) \|_1
	\label{eq:train_K}
\end{equation}
where $\mathbf{z}^{(i)}$ are true results for the KPIs from $\mathcal{D}$. On the other hand, in our hybrid approach, we use the meta-model only to approximate the computational expensive function $\mathbf{F}:\mathbf{p}\to\mathbf{y}$ for the FE calculation. Similar as before this approach can be written as
\begin{equation}
	\mathbf{z}\approx\mathbf{K}\bigl(\mathbf{p},\hat{\mathbf{F}}_\theta(\mathbf{p})\bigr)
	\label{eq:approx_F}
\end{equation}
where now $\theta$ represents the model parameters. The training loss function is now
\begin{equation}
	\min_\theta {L}(\theta) :=
	\sum_i \|\mathbf{y}^{(i)} - \hat{\mathbf{F}}_{\theta}(\mathbf{p}^{(i)}) \|_1
	\label{eq:train_F}
\end{equation}
where $\mathbf{y}^{(i)}$ are true results for the intermediate measures from~$\mathcal{D}$.

\subsection{Network architecture and training}
\begin{figure*}
	\centering
	\input{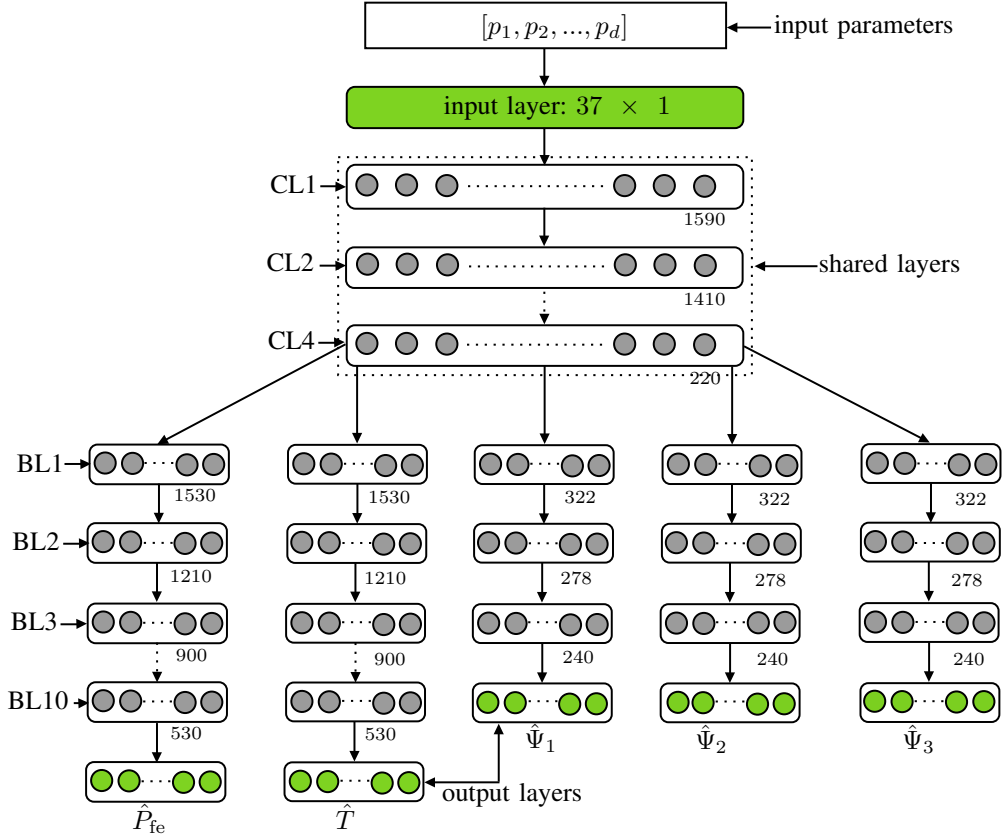}
	\vspace{-0.5em}
	\caption{Proposed multi-branch DNN network.}
	\label{fig:MDNN}
\end{figure*}
The performance of deep learning models depends strongly on the choice of hyperparameters. They are fixed before the training, whereas the model parameters, such as biases, weights of DNN are optimized during training. For better understanding, we divide the hyperparameters into two categories: model hyperparameters and learning hyperparameters. The model hyperparameters are specified as the width and height of different network layers, filter size, stride, type of layer (dense, convolutional, max-pooling, etc.), and so on. The learning hyperparameters include, for example, learning rate, activation functions, optimizer, loss function, batch size, epochs, etc. We propose the multi-branch DNN as depicted in \autoref{fig:MDNN}. Initially, this structure was obtained by trial and error, i.e., evaluating more than ten different configurations. The hyperparameters were then numerically optimized. The parameter space is given in~\autoref{tab:HPO}. The hyperparameter optimization (HPO) was performed using an in-house optimization tool that includes implementation of the Asynchronous Successive Halving Algorithm algorithm \cite{MLSYS2020_f4b9ec30} for the multi-objective case. This tool has evaluated 800 different configurations within search space defined in \autoref{tab:HPO}. The entire HPO process took approximately two days with parallelization across four GPUs. The final hyperparameter configuration is given in the last column of \autoref{tab:HPO}. As shown in \autoref{fig:MDNN}, the input layer has a size of $37\times 1$ as there are 37 varying input parameters. There are four common layers (CL: $1590 \rightarrow 1410 \rightarrow 810 \rightarrow 210$) to exploit correlation among all the output measures. There are total five distinct branches with different number of branch layers (BL). There are two identical branches for torque $\hat{T}$ and iron loss $\hat{P}_{\mathrm{fe}}$ prediction, and three equal size branches for flux $\hat{\Psi}$ prediction. The network layers for iron loss and torque are $1530 \rightarrow 1210 \rightarrow 900 \rightarrow 880 \rightarrow 750 \rightarrow 660 \rightarrow 610 \rightarrow 580 \rightarrow 550 \rightarrow 530$. The branch configuration for other three branches are $322 \rightarrow 278 \rightarrow 240$. 

As stated earlier in the section \autoref{sec:dataset}, each branch predicts intermediate measures for every operating point. In this study, we predict 15 time steps by the flux and torque branches, so the size of output layers remains $15 \times 1$ for both output measures. These figures, however, can be modified as needed. The output layer for iron losses is $4 \times 1$ in size. The total number of trainable network parameters is noted to be around 2.3 million. The actual number of machine designs ${N_{\mathrm{EM}}}= 44877$ is divided into training, validation and testing. The training sample consists $N_{\mathrm{train}}= 40390$ ($\sim 90\%$), whereas validation ${N_{\mathrm{val}}} = 2243$ and test samples ${N_{\mathrm{test}}} = 2244$ comprises $\sim 5\%$ each. Since each machine design is evaluated for all operating points, the total number of training samples is obtained after multiplication by~$N_{\mathrm{OP}}=37$.
    
\begin{figure}
	\centering
	\input{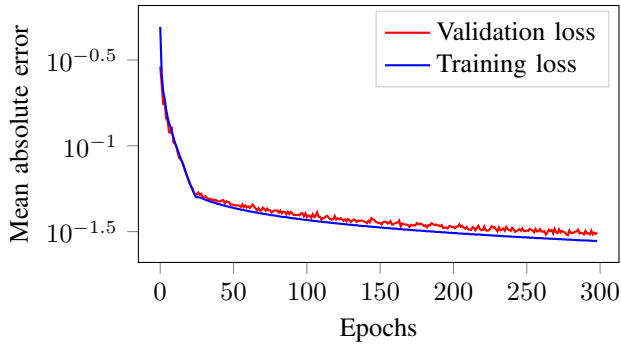}
	\vspace{-0.6em}
	\caption{Training curve.}
	\label{fig:TRN}
\end{figure}

The model training for the final configuration was carried out on a NVIDIA Quadro M2000M GPU using standard back-propagation \cite{rumelhart1986learning}. Tensorflow2 \cite{tensorflow2015-whitepaper}, a deep learning framework, was used to implement the training pipeline. It took around $\sim2$ hours to finish the training with early stopping criteria (10 epochs over validation error) for a maximum of 300 epochs. The training and validation curve is illustrated in \autoref{fig:TRN}. To allow a fair comparison with the direct DL approach, we used the same network architecture and model hyperparameters from \cite{9333549}. The model learning hyperparameters are the same for both approaches.

\begin{table*}
	\caption{Details of Hyperparameters}
	\label{tab:HPO}
	\begin{tabularx}{\linewidth}{|X|r|r|r|}
	\hline
	\textbf{Hyperparameter}                    & \textbf{Min}             & \textbf{Max}                                  & \textbf{Optimized values}    \\ \hline
	Learning rate                              & \hspace{2cm}$10^{-6}$                & $10^{-3}$                                     & $2.6\cdot10^{-4}$            \\ \hline
	Average number of neurons per hidden layer & $50\phantom{^{-6}}$      & $1000\phantom{^{-3}}$                         & $991$                        \\ \hline
	Number of common layers                    & $1\phantom{^{-6}}$       & $6\phantom{^{-3}}$                            & $4$                          \\ \hline
	Number of branch layers                    & $3\phantom{^{-6}}$       & $12\phantom{^{-3}}$                           & $10$                         \\ \hline
	Batchsize                                  & $80\phantom{^{-6}}$      & $150\phantom{^{-3}}$                          & $132$                        \\ \hline
	Activation functions                       & \multicolumn{2}{c|}{relu, elu, tanh, softplus}                           & elu                          \\ \hline
	Optimizers                                 & \multicolumn{2}{c|}{Adam, adamax, adagrad, nadam \cite{8769211}}         & Adam                         \\ \hline
	Loss functions                             & \multicolumn{2}{c|}{Mean absolute error, mean squared error, huber loss} & Mean absolute error          \\ \hline
	\end{tabularx}%
\end{table*} 

\section{Results and analysis}\label{sec:res}
We first showcase the evaluation of the intermediate measures using classical machine learning and multi-branch DNN. Following that, we explain the quantitative study with empirical results of the hybrid and direct DL approach. The characterization of the performance of a PMSM is a non-linear multi-output regression problem, and thus we chose to evaluate the performance of meta-models using several evaluation metrics, i.e., the mean relative error (MRE), mean absolute error (MAE) and the Pearson correlation coefficient (PCC).

\subsection{Discussion of the intermediate measures prediction}
\begin{figure}[!t]
	\centering
	\includegraphics[width = 0.9\columnwidth]{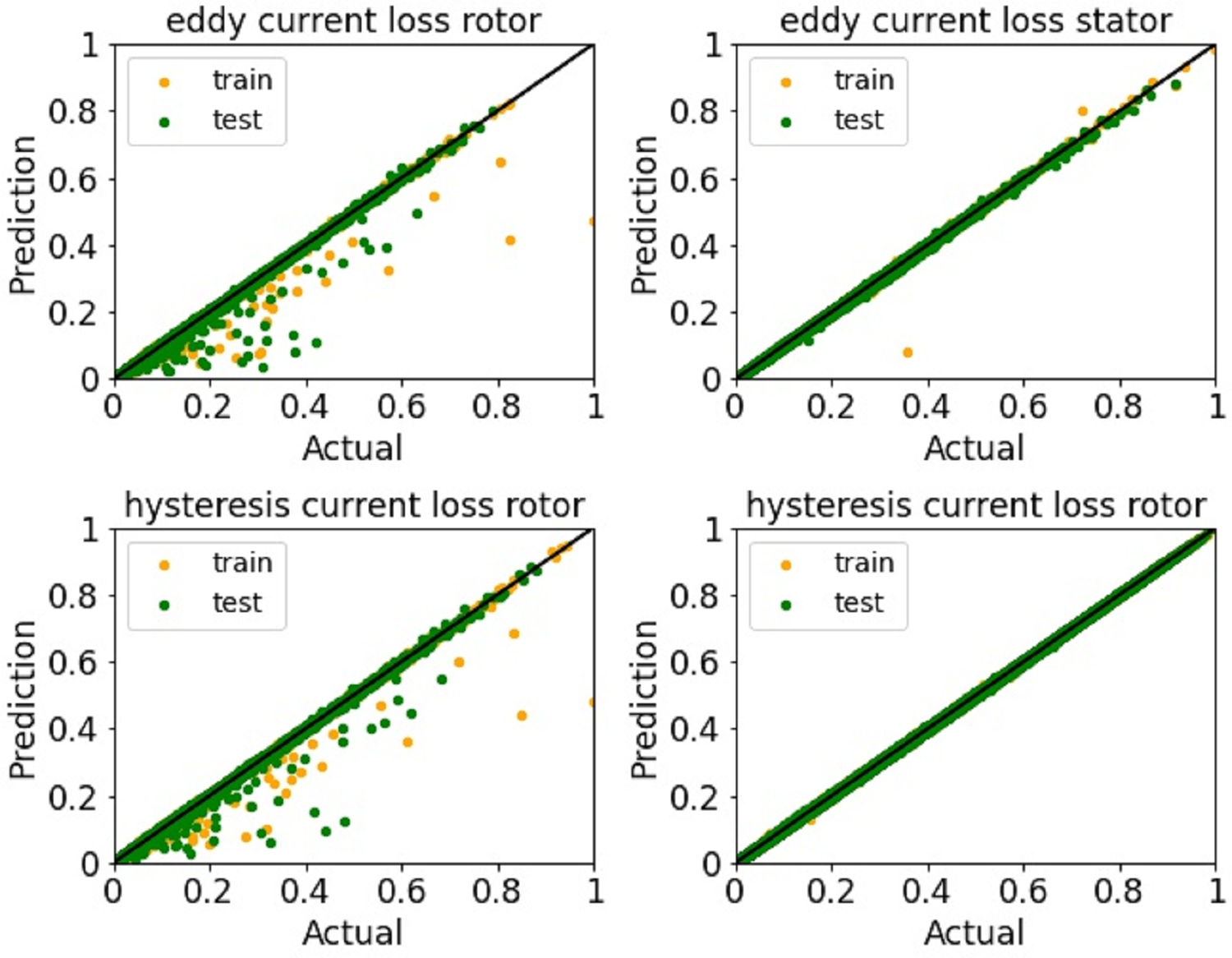}
	\vspace{-0.6em}
	\caption{Prediction plot of losses over test samples.}\label{fig:Iron_loss}
\end{figure}

\begin{table}
	\caption{Intermediate measures over test samples
	}
	\label{tab:IMFT}
	\begin{tabularx}{\linewidth}{|X|r|r|r|r|}
		\hline
		\multirow{2}{*}{\textbf{Measure}}    & \multicolumn{2}{c|}{\textbf{Multi-branch DNN}}
		                                     & \multicolumn{2}{c|}{\textbf{Random forest}}
		                                     \\ \cline{2-5} 
		                                     & \multicolumn{1}{c|}{$\varepsilon_{\textrm{MRE}}$}
		                                     & \multicolumn{1}{c|}{$\varepsilon_{\textrm{PCC}}$}
		                                     & \multicolumn{1}{c|}{$\varepsilon_{\textrm{MRE}}$}
		                                     & \multicolumn{1}{c|}{$\varepsilon_{\textrm{PCC}}$}
		                                     \\ \hline
		eddy current loss (rotor)      & $1.5\cdot10^{0\phantom{-}}$     & $0.98$         & $9.02\cdot10^{0\phantom{-}}$   & $0.97$         \\ \hline
		eddy current loss (stator)     & $5.9\cdot10^{-1}$               & $0.99$         & $7.32\cdot10^{0\phantom{-}}$   & $0.98$         \\ \hline
		hysteresis loss (rotor)        & $1.6\cdot10^{0\phantom{-}}$     & $0.98$         & $9.48\cdot10^{0\phantom{-}}$   & $0.97$         \\ \hline
		hysteresis loss (stator)       & $4.8\cdot10^{-1}$               & $0.99$         & $7.09\cdot10^{0\phantom{-}}$   & $0.98$         \\ \hline
		\hline
		 & \multicolumn{1}{c|}{$\varepsilon_{\textrm{MAE}}$} & \multicolumn{1}{c|}{$\varepsilon_{\textrm{PCC}}$} & \multicolumn{1}{c|}{$\varepsilon_{\textrm{MAE}}$}   & \multicolumn{1}{c|}{$\varepsilon_{\textrm{PCC}}$}         \\ \hline
		\hline
		Electromagnetic\;torque\;$T$   & $6.5\cdot10^{-1}$               & $0.99$         & $8.54\cdot10^{0\phantom{-}}$   & $0.97$         \\ \hline
		Flux linkage $\Psi_{1}$ coil 1 & $1.0\cdot10^{-6}$               & $0.99$         & $2.40\cdot10^{-3}$             & $0.98$         \\ \hline
		Flux linkage $\Psi_{2}$ coil 2 & $1.0\cdot10^{-5}$               & $0.99$         & $2.30\cdot10^{-3}$             & $0.98$         \\ \hline
		Flux linkage $\Psi_{3}$ coil 3 & $1.0\cdot10^{-6}$               & $0.99$         & $2.80\cdot10^{-3}$             & $0.98$         \\ \hline
	\end{tabularx}%
\end{table}
The problem at hand requires multi-output regression and there are several algorithms and implementations available. We considered several state-of-the-art approaches using the sci-kit-learn library \cite{JMLR:v12:pedregosa11a,buitinck:hal-00856511} with their default settings. The library contains an implementation of K-Nearest Neighbour (with KNeighborsRegressor) \cite{Omohundro89fiveballtree,10.1145/361002.361007}, Support Vector Regression \cite{10.1145/1961189.1961199}, and Gaussian Process Regression \cite{C.E.Rasmussen}, which were unable to process the data due to memory constraints. These algorithms require higher computational power (ca. five times higher in our experiments) as compared to our existing resources (NVIDIA Quadro M2000M GPU) to process such a large volume of high dimensional data. Random forest \cite{breiman2001random} is an ensemble method, that combines multiple decision trees to train a meta-model. The final model makes predictions by taking the average of the results of the different trees (bootstrap aggregation). The algorithm succeeded to train the meta-model using the default settings (number of estimators is $100$, the evaluation criterion is squared error, the minimum number of samples for each split is $2$, the minimum samples leaf is $1$ and the bootstrap aggregation enabled). However, with low accuracy and high computational costs: the training time is almost $\sim 10$ higher than for the multi-branch DNN, and the final test accuracy is much lower compared to multi-branch DNN. The comparison between multi-branch DNN and random forest over test samples for intermediate measures is given in \autoref{tab:IMFT}. From the results, it is evident that the multi-branch DNN outperforms the prediction accuracy of random forest. We use the MRE to evaluate iron losses because its range varies largely, while torque and flux with MAE. The reason is that the torque $T$ and the fluxes $\Psi_{1}, \Psi_{2}, \Psi_{3}$ are more sensitive. For the testing, the MAE is evaluated over the mean of the predicted time steps for each operating point. \autoref{fig:Iron_loss} depicts the prediction curve over each operating point for all the test samples. It is observed that the multi-branch DNN predicts intermediate measures with high accuracy close to the ground truth for all the test samples. The evaluation time for new machine designs is $\sim100$ ms/sample, which is much lower in comparison to the actual FE simulation ($\sim3-5$ hours/sample). \autoref{fig:evaloppts} illustrates two operating points for different input electrical excitation conditions for one test machine design. Figure~\ref{fig:op1p} unveils that the multi-branch DNN has poor prediction accuracy for the operating point at zero input phase current and open circuit (no load) condition. 

\subsection{Quantitative analysis}
\begin{figure*}[ht]
    \centering
        \subfloat[Operating point: zero current $I$ and $\alpha = 0\degree$]{\input{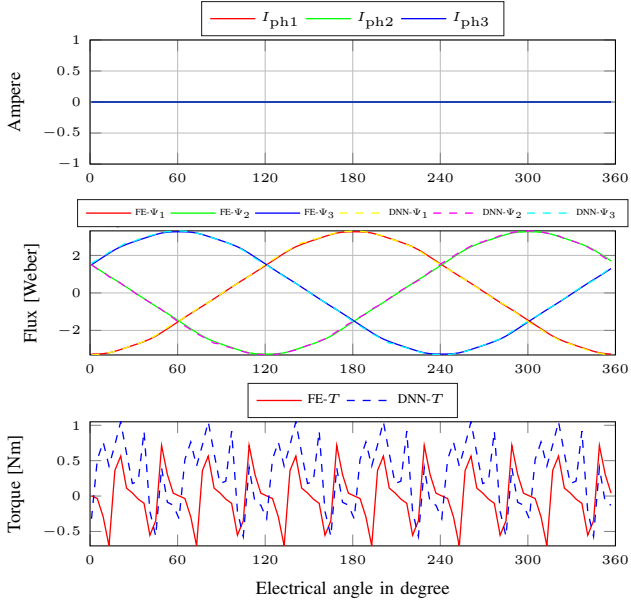}%
        \label{fig:op1p}}
        \hfil
        \subfloat[Operating point: maximal current $I$ and $\alpha = 0\degree$]{\input{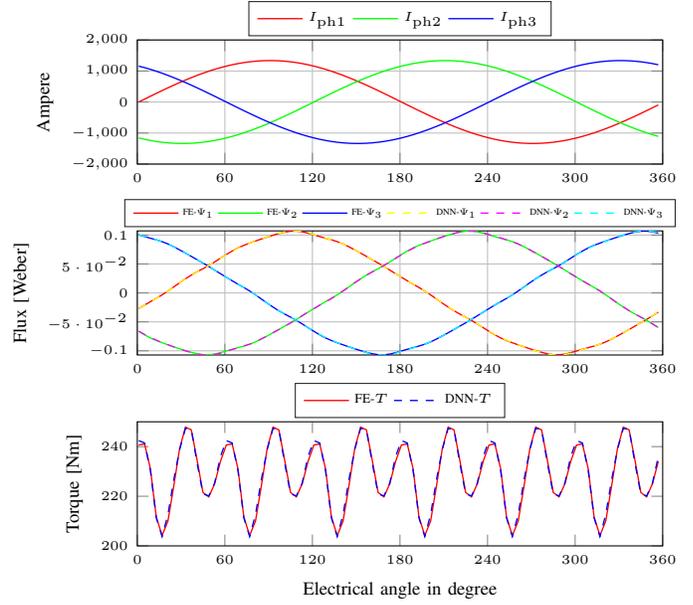}%
        \label{fig:op2p}}
    \caption{Operating points for the test design.}
    \label{fig:evaloppts}
\end{figure*}
\begin{figure}
	\centering
	\begin{tikzpicture}
\pgfplotsset{every axis title/.append style={at={(0.5,0.85)}}}
\tikzstyle{every node}=[font=\tiny]
\begin{axis}[%
width=1.1in,
height=0.65in,
at={(2.08in,1.1in)},
scale only axis,
xmin=0,
xmax=22002,
xlabel style={font=\scriptsize},
xlabel={Speed (1/min)},
ymin=0,
ymax=800,
ylabel style={font=\scriptsize},
ylabel={Torque [Nm]},
axis background/.style={fill=white},
title style={font=\scriptsize},
title={Maximum torque curve},
legend style={legend cell align=left, align=left, draw=white!15!black,nodes={scale=0.5, transform shape}}
]
\addplot [color=blue, dashed, line width=0.5pt]
  table[row sep=crcr]{%
1	710.418041735044\\
1001	709.682199230951\\
2001	708.970063795074\\
3001	708.263430375162\\
4001	707.559872106693\\
5001	695.713512108743\\
6001	587.726323449979\\
7001	471.299821002191\\
8001	385.682317507922\\
9001	324.791201507317\\
10001	279.889255187089\\
11001	245.921690513273\\
12001	219.236803016556\\
13001	197.432743061036\\
14001	179.827088214665\\
15001	164.724617780564\\
16001	152.271187767471\\
17001	141.137114733815\\
18001	131.645359286882\\
19001	123.272967305549\\
20001	115.903366181283\\
};
\addlegendentry{DNN}

\addplot [color=red, line width=0.5pt]
  table[row sep=crcr]{%
1	711.550907677214\\
1001	710.817890660167\\
2001	710.108447863833\\
3001	709.404507083466\\
4001	708.70364145454\\
5001	696.363925608543\\
6001	585.858177774796\\
7001	468.614406859927\\
8001	384.073236025721\\
9001	323.788008827475\\
10001	278.949384741413\\
11001	244.501271446401\\
12001	217.706057244876\\
13001	195.753757877027\\
14001	177.709635790131\\
15001	162.755219920336\\
16001	150.000777190838\\
17001	138.936620657036\\
18001	129.356716135277\\
19001	120.914702257706\\
20001	113.533512919013\\
};
\addlegendentry{FE SIMULATION}

\end{axis}

\begin{axis}[%
width=1.1in,
height=0.65in,
at={(3.75in,1.1in)},
scale only axis,
xmin=0,
xmax=22002,
xlabel style={font=\scriptsize},
xlabel={Speed (1/min)},
ymin=0,
ymax=400,
ylabel style={font=\scriptsize},
ylabel={Shaft power [KW]},
axis background/.style={fill=white},
title style={font=\scriptsize},
title={Machine shaft power curve},
legend style={at={(0.97,0.03)},anchor=south east,legend cell align=left, align=left, draw=white!15!black,nodes={scale=0.5, transform shape}}
]
\addplot [color=blue, dashed, line width=0.5pt]
  table[row sep=crcr]{%
1	0.0743948033630821\\
1001	74.3920638623631\\
2001	148.560586107047\\
3001	222.581688140281\\
4001	296.456081654582\\
5001	364.347598055952\\
6001	369.340886571027\\
7001	345.530167304866\\
8001	323.148851305144\\
9001	306.142507836883\\
10001	293.128652572423\\
11001	283.307204159999\\
12001	275.524050325655\\
13001	268.797085685913\\
14001	263.659112432738\\
15001	258.766074464039\\
16001	255.148769718124\\
17001	251.27212942884\\
18001	248.159442038702\\
19001	245.286093148056\\
20001	242.759579853089\\
};
\addlegendentry{DNN}

\addplot [color=red, line width=0.5pt]
  table[row sep=crcr]{%
1	0.0745134368071295\\
1001	74.5111121200504\\
2001	148.799127920173\\
3001	222.940287453953\\
4001	296.935302413908\\
5001	364.688221879222\\
6001	368.166900393472\\
7001	343.561374709345\\
8001	321.800661852255\\
9001	305.196915956854\\
10001	292.144324120266\\
11001	281.670850108711\\
12001	273.600298158004\\
13001	266.511211937849\\
14001	260.554543302392\\
15001	255.672345304333\\
16001	251.344422527587\\
17001	247.354500578977\\
18001	243.845211665489\\
19001	240.593664363089\\
20001	237.795922616746\\
};
\addlegendentry{FE SIMULATION}

\end{axis}

\begin{axis}[%
width=1.1in,
height=0.65in,
at={(2.08in,0in)},
scale only axis,
xmin=0,
xmax=22002,
xlabel style={font=\scriptsize},
xlabel={Speed (1/min)},
ymin=0,
ymax=400,
ylabel style={font=\scriptsize},
ylabel={Open circuit voltage [V]},
axis background/.style={fill=white},
title style={font=\scriptsize},
title={Open circuit voltage curve},
legend style={at={(0.97,0.03)}, anchor=south east, legend cell align=left, align=left, draw=white!15!black,nodes={scale=0.5, transform shape}}
]
\addplot [color=blue, dashed, line width=0.5pt]
  table[row sep=crcr]{%
1	0.0186743816690835\\
1001	18.6930560507526\\
2001	37.3674377198361\\
3001	56.0418193889196\\
4001	74.7162010580031\\
5001	93.3905827270866\\
6001	112.06496439617\\
7001	130.739346065254\\
8001	149.413727734337\\
9001	168.088109403421\\
10001	186.762491072504\\
11001	205.436872741588\\
12001	224.111254410671\\
13001	242.785636079755\\
14001	261.460017748838\\
15001	280.134399417922\\
16001	298.808781087005\\
17001	317.483162756089\\
18001	336.157544425172\\
19001	354.831926094256\\
20001	373.506307763339\\
};
\addlegendentry{DNN}

\addplot [color=red, line width=0.5pt]
  table[row sep=crcr]{%
1	0.0185715341865092\\
1001	18.5901057206957\\
2001	37.1616399072048\\
3001	55.733174093714\\
4001	74.3047082802232\\
5001	92.8762424667323\\
6001	111.447776653241\\
7001	130.019310839751\\
8001	148.59084502626\\
9001	167.162379212769\\
10001	185.733913399278\\
11001	204.305447585787\\
12001	222.876981772296\\
13001	241.448515958806\\
14001	260.020050145315\\
15001	278.591584331824\\
16001	297.163118518333\\
17001	315.734652704842\\
18001	334.306186891351\\
19001	352.877721077861\\
20001	371.44925526437\\
};
\addlegendentry{FE SIMULATION}

\end{axis}

\begin{axis}[%
width=1.1in,
height=0.65in,
at={(3.75in,0in)},
scale only axis,
xmin=0,
xmax=22002,
xlabel style={font=\scriptsize},
xlabel={Speed (1/min)},
ymin=0,
ymax=400,
ylabel style={font=\scriptsize},
ylabel={Short circuit current [A]},
axis background/.style={fill=white},
title style={font=\scriptsize},
title={Short circuit current curve},
legend style={at={(0.97,0.03)}, anchor=south east, legend cell align=left, align=left, draw=white!15!black,nodes={scale=0.5, transform shape}}
]
\addplot [color=blue, dashed, line width=0.5pt]
  table[row sep=crcr]{%
1	0.945582592838878\\
1001	303.513182874455\\
2001	319.588034043284\\
3001	322.427099109108\\
4001	324.319552410537\\
5001	324.32025353962\\
6001	325.266321912677\\
7001	325.26660517345\\
8001	325.266777304569\\
9001	325.266934521626\\
10001	325.267007113601\\
11001	325.26707546473\\
12001	325.267139400288\\
13001	325.267198743666\\
14001	325.26725331637\\
15001	325.26725331637\\
16001	325.26730293803\\
17001	325.26730293803\\
18001	325.267347426401\\
19001	325.267347426401\\
20001	325.267347426401\\
};
\addlegendentry{DNN}

\addplot [color=red, line width=0.5pt]
  table[row sep=crcr]{%
1	0.945983498266219\\
1001	300.804164626523\\
2001	316.885749243004\\
3001	319.725996137186\\
4001	320.673300734169\\
5001	321.619934965809\\
6001	321.62035798921\\
7001	321.620647186485\\
8001	322.566851453091\\
9001	322.567007365144\\
10001	322.567079354559\\
11001	322.567147138328\\
12001	322.56721054318\\
13001	322.567269393969\\
14001	322.567323513684\\
15001	322.567323513684\\
16001	322.567372723452\\
17001	322.567372723452\\
18001	322.567416842541\\
19001	322.567416842541\\
20001	322.567416842541\\
};
\addlegendentry{FE SIMULATION}

\end{axis}
\end{tikzpicture}%
	\caption{Different performance curves for one test design.
	}
	\label{fig:PC}
\end{figure}
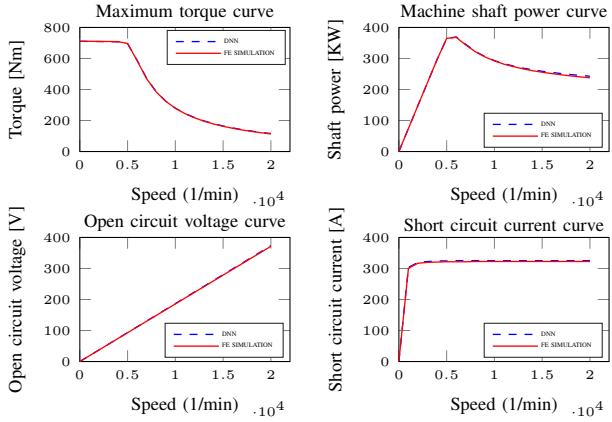
\begin{figure*}[!ht]
    \centering
        \subfloat[Classical approach]{\includegraphics[width=0.33\textwidth]{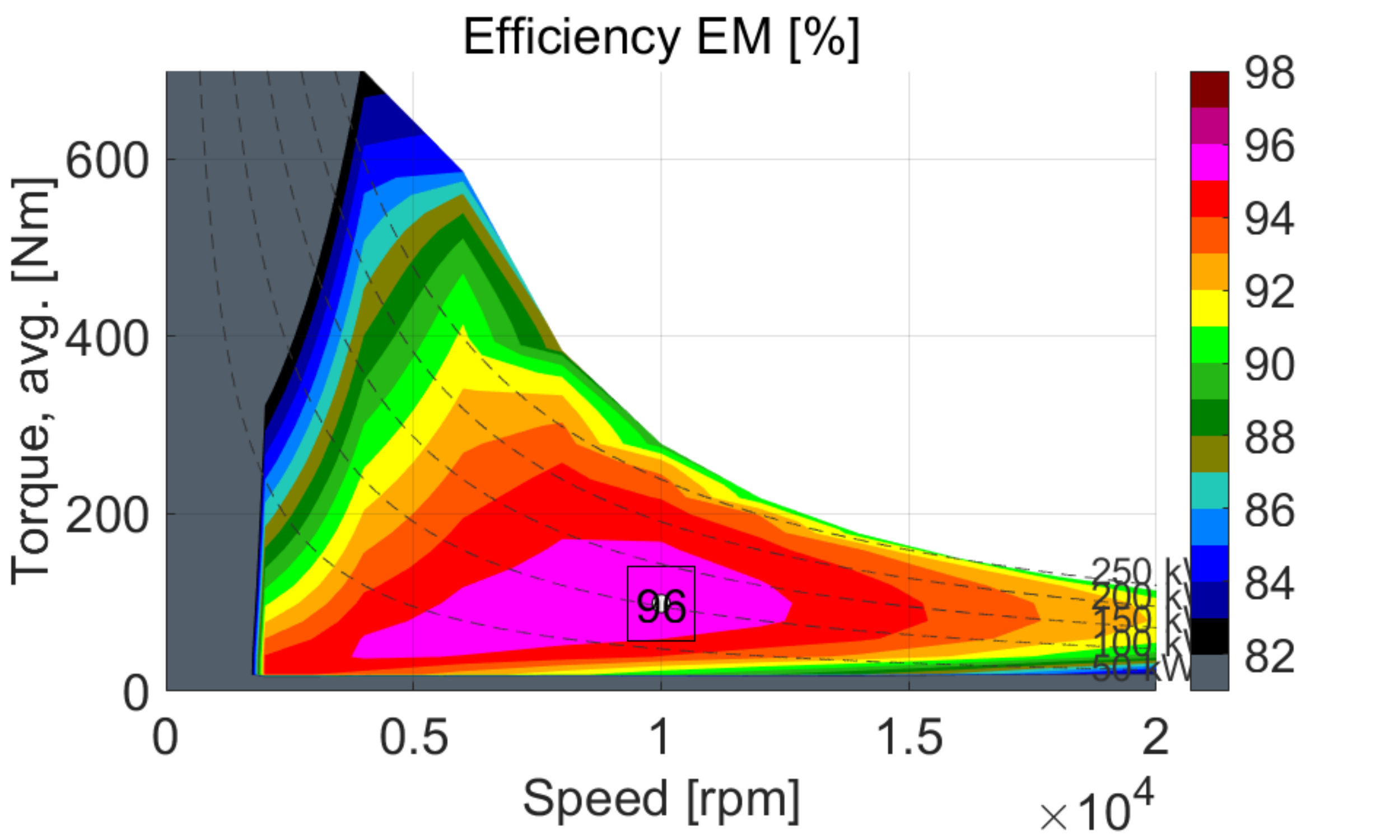}%
        \label{fig:emorg}}
        \hfil
        \subfloat[Hybrid approach]{\includegraphics[width=0.33\textwidth]{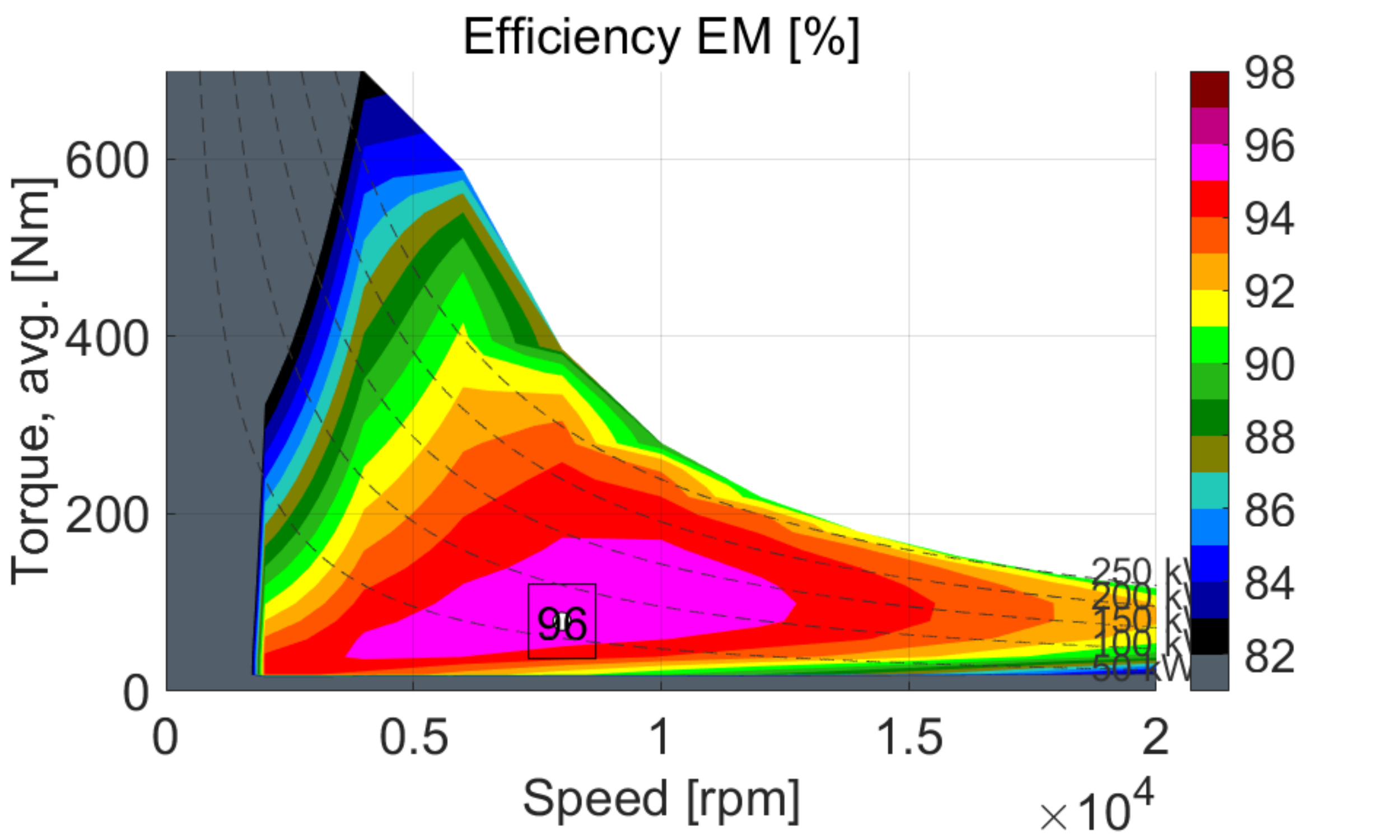}%
        \label{fig:empred}}
        \hfil
        \subfloat[Absolute calculation difference]{\includegraphics[width=0.33\textwidth]{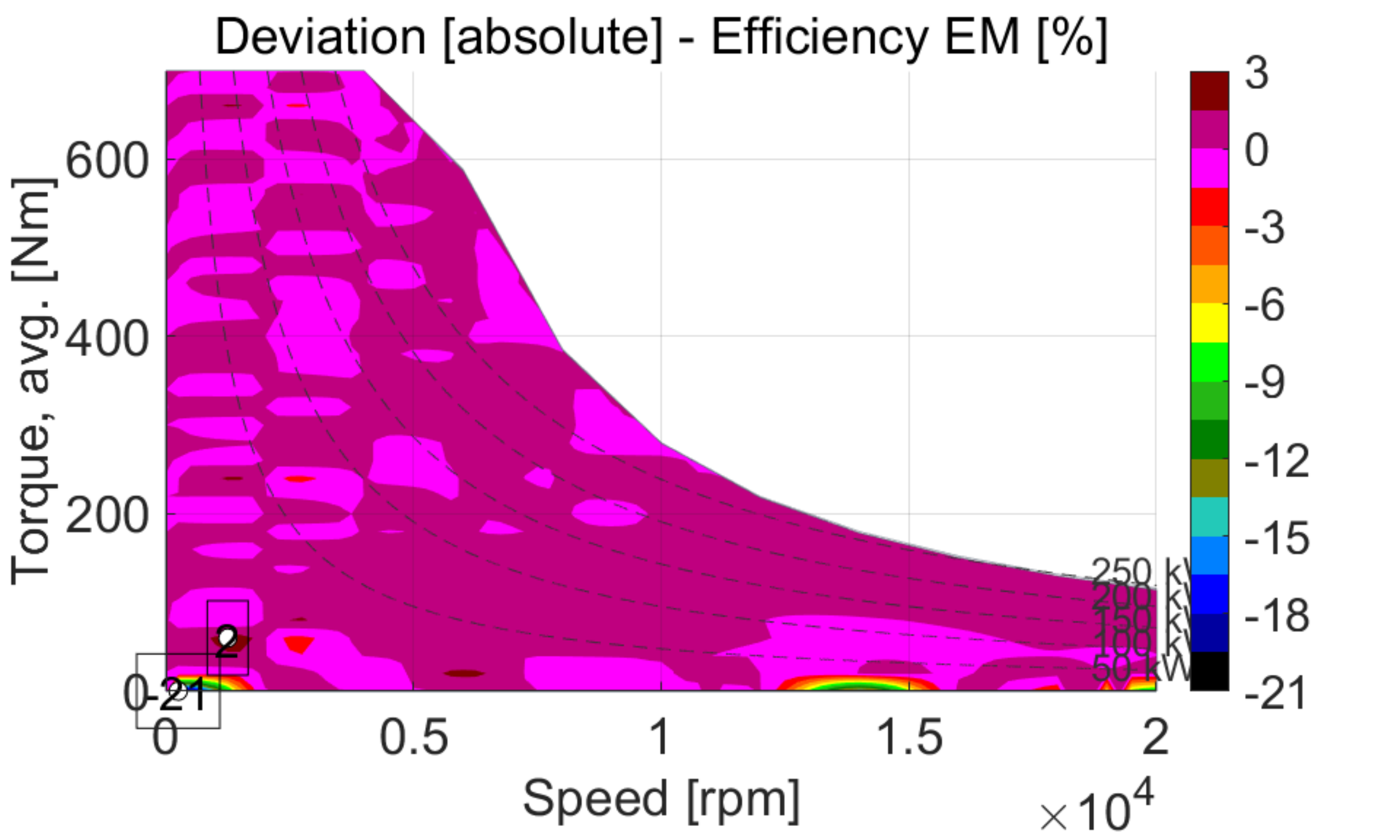}%
        \label{fig:emdiff}}
    \caption{Illustration of efficiency map calculation.
    }
    \label{fig:EM_map}
\end{figure*}

\begin{figure}[ht]
	\centering
	\includegraphics[width=0.95\linewidth]{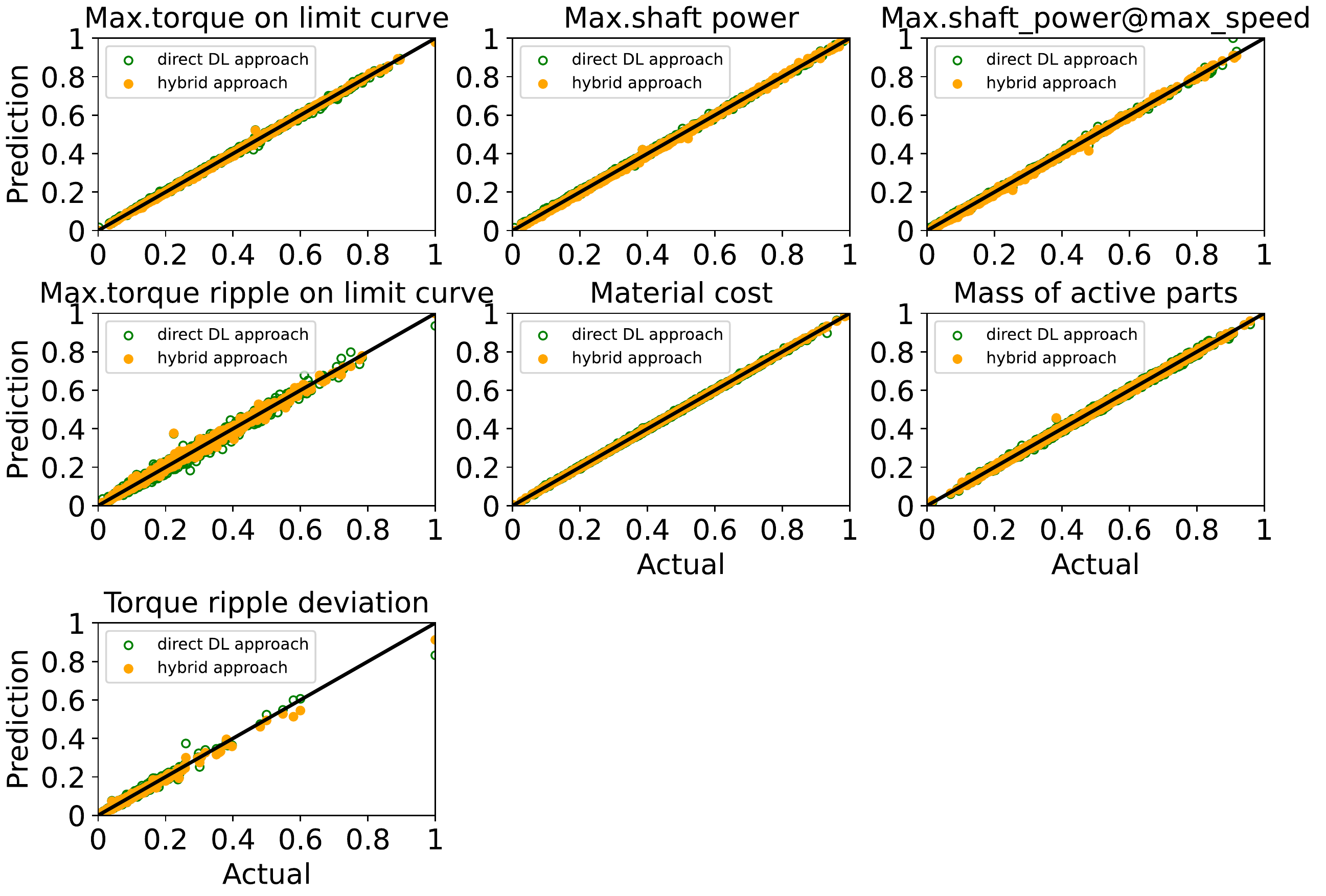}
	\caption{KPIs prediction plot over test samples.}
	\label{fig:pred}
\end{figure}

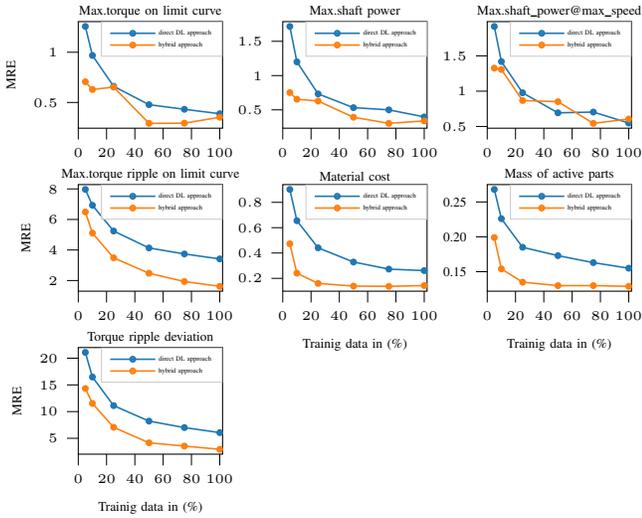
\begin{figure}[!t]
	\centering
\begin{tikzpicture}
\pgfplotsset{every axis title/.append style={at={(0.5,0.8)}}}
\tikzstyle{every node}=[font=\tiny]
\definecolor{color0}{rgb}{0.12156862745098,0.466666666666667,0.705882352941177}
\definecolor{color1}{rgb}{1,0.498039215686275,0.0549019607843137}

\begin{groupplot}[
        group style={group size=3 by 3,
            horizontal sep = 0.8 cm, 
            vertical sep = 0.75 cm}, 
        width = 3.5 cm, 
        height = 3 cm, 
        ]
\nextgroupplot[
legend cell align={left},
legend style={fill opacity=0.8, draw opacity = 1, text opacity=1, at = {(1,1)}, draw=white!80!black,nodes={scale=0.5, transform shape}},
tick align=outside,
tick pos=left,
title={Max.torque on limit curve},
x grid style={white!69.0196078431373!black},
xmin=0, xmax=102,
xtick style={color=black},
y grid style={white!69.0196078431373!black},
ylabel={MRE},
ymin=0.24375, ymax=1.30525,
ytick style={color=black}
]
\addplot [semithick, color0, mark=*, mark size=1, mark options={solid}]
table {%
5 1.257
10 0.968
25 0.661
50 0.478
75 0.432
100 0.388
};
\addlegendentry{direct DL approach}
\addplot [semithick, color1, mark=*, mark size=1, mark options={solid}]
table {%
5 0.706
10 0.629
25 0.653
50 0.292
75 0.294
100 0.352
};
\addlegendentry{hybrid approach}

\nextgroupplot[
legend cell align={left},
legend style={fill opacity=0.8, draw opacity = 1, text opacity=1, at = {(1,1)}, draw=white!80!black,nodes={scale=0.5, transform shape}},
tick align=outside,
tick pos=left,
title= {Max.shaft power},
x grid style={white!69.0196078431373!black},
xmin=0, xmax=102,
xtick style={color=black},
y grid style={white!69.0196078431373!black},
ymin=0.2323, ymax=1.7877,
ytick style={color=black}
]
\addplot [semithick, color0, mark=*, mark size=1, mark options={solid}]
table {%
5 1.717
10 1.2
25 0.733
50 0.532
75 0.5
100 0.397
};
\addlegendentry{direct DL approach}
\addplot [semithick, color1, mark=*, mark size=1, mark options={solid}]
table {%
5 0.752
10 0.655
25 0.629
50 0.392
75 0.303
100 0.338
};
\addlegendentry{hybrid approach}

\nextgroupplot[
legend cell align={left},
legend style={fill opacity=0.8, draw opacity = 1, text opacity=1, at = {(1,1)}, draw=white!80!black,nodes={scale=0.5, transform shape}},
tick align=outside,
tick pos=left,
title={Max.shaft\_power@max\_speed},
x grid style={white!69.0196078431373!black},
xmin=0, xmax=102,
xtick style={color=black},
y grid style={white!69.0196078431373!black},
ymin=0.47535, ymax=1.98565,
ytick style={color=black}
]
\addplot [semithick, color0, mark=*, mark size=1, mark options={solid}]
table {%
5 1.917
10 1.419
25 0.977
50 0.693
75 0.704
100 0.549
};
\addlegendentry{direct DL approach}
\addplot [semithick, color1, mark=*, mark size=1, mark options={solid}]
table {%
5 1.326
10 1.307
25 0.867
50 0.851
75 0.544
100 0.605
};
\addlegendentry{hybrid approach}

\nextgroupplot[
legend cell align={left},
legend style={fill opacity=0.8, draw opacity = 1, text opacity=1, at = {(1,1)}, draw=white!80!black,nodes={scale=0.5, transform shape}},
tick align=outside,
tick pos=left,
title={Max.torque ripple on limit curve},
x grid style={white!69.0196078431373!black},
xmin=0, xmax=102,
xtick style={color=black},
y grid style={white!69.0196078431373!black},
ylabel={MRE},
ymin=1.30265, ymax=8.28435,
ytick style={color=black}
]
\addplot [semithick, color0, mark=*, mark size=1, mark options={solid}]
table {%
5 7.967
10 6.928
25 5.241
50 4.134
75 3.74
100 3.416
};
\addlegendentry{direct DL approach}
\addplot [semithick, color1, mark=*, mark size=1, mark options={solid}]
table {%
5 6.49
10 5.109
25 3.484
50 2.476
75 1.934
100 1.62
};
\addlegendentry{hybrid approach}

\nextgroupplot[
legend cell align={left},
legend style={fill opacity=0.8, draw opacity = 1, text opacity=1, at = {(1,1)}, draw=white!80!black,nodes={scale=0.5, transform shape}},
tick align=outside,
tick pos=left,
title={Material cost},
x grid style={white!69.0196078431373!black},
xlabel={Trainig data in (\%)},
xmin=0, xmax=102,
xtick style={color=black},
y grid style={white!69.0196078431373!black},
ymin=0.09875, ymax=0.94025,
ytick style={color=black}
]
\addplot [semithick, color0, mark=*, mark size=1, mark options={solid}]
table {%
5 0.902
10 0.655
25 0.441
50 0.329
75 0.273
100 0.261
};
\addlegendentry{direct DL approach}
\addplot [semithick, color1, mark=*, mark size=1, mark options={solid}]
table {%
5 0.473
10 0.241
25 0.16
50 0.139
75 0.137
100 0.143
};
\addlegendentry{hybrid approach}

\nextgroupplot[
legend cell align={left},
legend style={fill opacity=0.8, draw opacity = 1, text opacity=1, at = {(1,1)}, draw=white!80!black,nodes={scale=0.5, transform shape}},
tick align=outside,
tick pos=left,
title={Mass of active parts},
x grid style={white!69.0196078431373!black},
xlabel={Trainig data in (\%)},
xmin=0, xmax=102,
xtick style={color=black},
y grid style={white!69.0196078431373!black},
ymin=0.12205, ymax=0.27495,
ytick style={color=black},
ytick={0.1,0.15,0.2,0.25,0.3},
yticklabels={0.10,0.15,0.20,0.25,0.30}
]
\addplot [semithick, color0, mark=*, mark size=1, mark options={solid}]
table {%
5 0.268
10 0.226
25 0.185
50 0.173
75 0.163
100 0.155
};
\addlegendentry{direct DL approach}
\addplot [semithick, color1, mark=*, mark size=1, mark options={solid}]
table {%
5 0.199
10 0.154
25 0.135
50 0.13
75 0.13
100 0.129
};
\addlegendentry{hybrid approach}

\nextgroupplot[
legend cell align={left},
legend style={fill opacity=0.8, draw opacity = 1, text opacity=1, at = {(1,1)}, draw=white!80!black,nodes={scale=0.5, transform shape}},
tick align=outside,
tick pos=left,
title={Torque ripple deviation},
x grid style={white!69.0196078431373!black},
xlabel={Trainig data in (\%)},
xmin=0, xmax=102,
xtick style={color=black},
y grid style={white!69.0196078431373!black},
ylabel={MRE},
ymin=2.02615, ymax=21.99885,
ytick style={color=black}
]
\addplot [semithick, color0, mark=*, mark size=1, mark options={solid}]
table {%
5 21.091
10 16.476
25 11.121
50 8.216
75 7.004
100 6.06
};
\addlegendentry{direct DL approach}
\addplot [semithick, color1, mark=*, mark size=1, mark options={solid}]
table {%
5 14.333
10 11.536
25 7.06
50 4.15
75 3.539
100 2.934
};
\addlegendentry{hybrid approach}

\nextgroupplot[
hide x axis,
hide y axis,
tick align=outside,
tick pos=left,
x grid style={white!69.0196078431373!black},
xmin=0, xmax=1,
xtick style={color=black},
y grid style={white!69.0196078431373!black},
ymin=0, ymax=1,
ytick style={color=black}
]
\end{groupplot}

 \end{tikzpicture}
	\caption{KPIs evaluation over varying training set size.}
	\label{fig:cmp}
\end{figure}

\begin{table}
	\caption{Hybrid and direct DL approach over test samples}
	\label{tab:cmphydl}
	\begin{tabularx}{\linewidth}{|X|r|r|r|r|}
		\hline
		\multirow{2}{*}{\textbf{KPIs}}                & \multicolumn{2}{c|}{\textbf{Multi-branch DNN}}
		                                             & \multicolumn{2}{c|}{\textbf{Direct DL approach}}
		                                             \\ \cline{2-5} 
		                                             & \multicolumn{1}{c|}{$\varepsilon_{\textrm{MRE}}$}
		                                             & \multicolumn{1}{c|}{$\varepsilon_{\textrm{PCC}}$}
		                                             & \multicolumn{1}{c|}{$\varepsilon_{\textrm{MRE}}$}
		                                             & \multicolumn{1}{c|}{$\varepsilon_{\textrm{PCC}}$}
		                                             \\ \hline
		$z_{1}$ & \hspace{1.5em}$0.35$ & \hspace{1.5em}$1.00$  &\hspace{1.5em} $0.39$ & \hspace{1.5em}$1.00$ \\ \hline
		$z_{2}$ & $0.34$ & $1.00$  & $0.40$ & $1.00$ \\ \hline
		$z_{3}$ & $0.60$ & $0.99$  & $0.55$ & $1.00$ \\ \hline
		$z_{4}$ & $1.62$ & $0.99$  & $3.42$ & $0.98$ \\ \hline
		$z_{5}$ & $0.14$ & $1.00$  & $0.26$ & $1.00$ \\ \hline
		$z_{6}$ & $0.13$ & $1.00$  & $0.16$ & $1.00$ \\ \hline
		$z_{7}$ & $2.93$ & $0.99$  & $6.06$ & $0.98$ \\ \hline
	\end{tabularx}%
\end{table}

Let us compare the proposed hybrid approach \autoref{fig:MTH} with the parameter-based direct DL approach described in \cite{9333549}. \autoref{fig:cmp} shows evaluation with MRE over an increasing training size from $5\%$ to the total training size. The hybrid approach consistently performs better for KPIs $z_{1}, z_{2}, z_{4}, z_{5},z_{6}, z_{7}$, while the direct DL approach is slightly more accurate for $z_{3}$. Both meta-models predict KPIs for unseen machine designs. \autoref{fig:pred} illustrates the prediction plot over the test samples for meta-models trained on the full training set. As explained in \autoref{fig:MTH}, the training of multi-branch DNN is independent of the system parameters and solely relies on varying geometry, electrical excitation, and material parameters, whereas in the direct DL approach, model training with output KPIs implicitly involve fixed value for the system parameters. This makes the hybrid approach more flexible than the direct DL approach. The post-processing of FE output only takes very little time ($\sim 3-5$ min.) and is performed using a physics-based in-house post-processing tool, and hence the hybrid approach opens up new possibilities for further analysis. Demonstrative examples for one test sample are shown in \autoref{fig:PC} and \autoref{fig:EM_map}. \autoref{fig:PC} displays different performance curves, e.g., maximum torque curve, open circuit, and short circuit voltage characteristic, and maximum shaft power at different rotor speeds. \autoref{fig:EM_map} shows the efficiency map for the given test design, see \cite{7310051} for a detailed interpretation. The Figure~\ref{fig:emorg} shows the efficiency map for the FE simulation, Figure~\ref{fig:empred} is calculated from the intermediate measures predicted with multi-branch DNN, while Figure~\ref{fig:emdiff} gives details on the deviation between the two. The difference is close to zero with a maximum near the low torque region ($\sim20\%$). A possible explanation is that multi-branch DNN does not predict well for the open circuit operation mode at zero input current, which is also reflected in Figure~\ref{fig:op1p} for the same associated sample. In the open circuit mode, the associated torque range for the electrical machine is low ($\sim 10^{-1}$ to $10^{-3}$ Nm). If the multi-branch DNN predicts values in the range of $\sim 10^{-2}$ to $10^{-4}$ Nm this results in a relatively large error. However, the relatively poor prediction in this region does not have significant impact on the calculation of the overall efficiency of the electrical machine at other operating points (especially high-efficiency operating point regions as shown in Figure~\ref{fig:emdiff}). 

There are also a few disadvantages of the proposed hybrid approach. The training time of the multi-branch DNN is roughly about $\sim2$ hours, which is about $6\times$ higher than the training time ($\sim20$ minutes) of the DNN defined for direct KPIs prediction \cite{9333549}. A possible explanation is that the hybrid approach must deal with a high number of samples for model training ($37\times$ compared to the direct DNN in this study), and also the number of model parameters is around $\sim2.3$ million. Therefore, the hybrid approach requires higher computational power compared to DNN for the direct KPIs prediction. Also, the time to estimate the KPIs in the hybrid approach is increased from milliseconds to seconds due to the post-processing tool. However, the computational time remains much lower than the time for a FE simulation, which takes around $\sim 3-5$ hours/sample on a single-core CPU.

\section{Conclusion}\label{sec:conclusion}
In this paper, we presented a physics and data-driven hybrid method for the performance analysis of the electrical machine. The dataset used for the research is generated from a real-world industry design workflow. The proposed multi-branch DNN is trained to predict the intermediate measures using supervised learning. This hybrid approach is better than the existing parameter-based direct DL approach in terms of KPIs estimation and flexibility. This gain can be explained by two facts. Firstly, learning a few time-steps of intermediate measures is expected to be simpler than multiple (possibly independent) cross-domain KPIs and secondly, the post-processing tool exploits the laws of physics and thus ensuring that the KPIs fulfill the right constraints. Also on the application side, this approach makes the calculation of KPIs independent from the system parameters by predicting intermediate measures during the optimization and it enables the analysis of electrical machines with the calculation of more complex performance measures, e.g. efficiency map and characteristic curves. We have demonstrated that the trained multi-branch DNN meta-model evaluates new designs at much lower computational effort than the FE simulation. In future work, the proposed hybrid approach can be applied to many query scenarios, e.g. uncertainty quantification, sensitivity analysis and multi-objective optimization. 


\bibliographystyle{IEEEtran}
\bibliography{main2}

\begin{IEEEbiography}[{\includegraphics[width=1in,height=1.25in,clip,keepaspectratio]{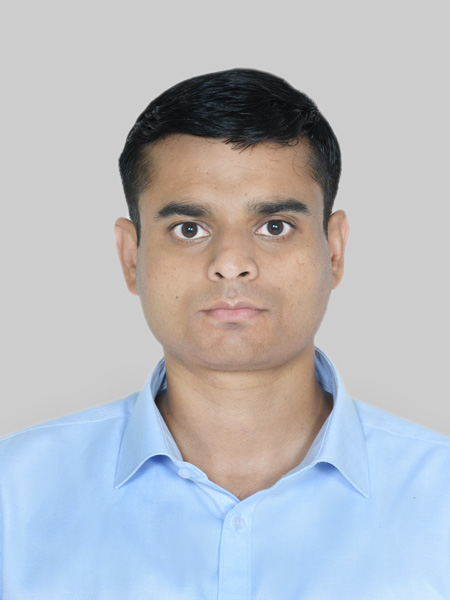}}]{Vivek Parekh} received his bachelor degree in electronics and communication from Gujarat technological university in 2012. He worked as an Assistant manager in Reliance Jio Infocomm Ltd. from 2013 to 2015 in operation and maintenance. He has obtained his M.Sc degree in information technology from the University of Stuttgart in 2019. Currently, he is pursuing a Ph.D. at TU Darmstadt in the Institut f\"{u}r Teilchenbeschleunigung und Elektromagnetische Felder under the fellowship of Robert Bosch GmbH. His area of interest comprises machine learning, deep learning, reinforcement learning, optimization of electrical machine design, and the industrial simulation process.
    \end{IEEEbiography}
    \begin{IEEEbiography}[{\includegraphics[width=1in,height=1.25in,clip,keepaspectratio]{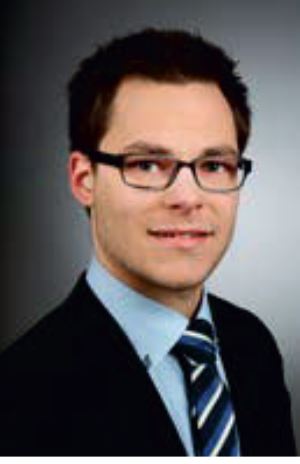}}]{Dominik Flore} received the bachelor in maschinenbau with specialization in mechatronik and master in maschinenbau with specialization in Product development from the University of Paderborn in 2012 and 2013, respectively. He obtained a Ph.D. degree with the topic "Experimentelle Untersuchung und Modellierung des Sch\"{a}digungsverhaltensfaserverst\"{a}rkter Kunststoffe" from ETH Z\"{u}rich, in 2016. Currently, he is working as a development engineer for the reliability of the electric machine at Robert Bosch GmbH, Stuttgart.Present research interests involve product development in the field of electrical machine with the application of machine learning, optimization of the industrial simulation process.
    \end{IEEEbiography}
    \begin{IEEEbiography}[{\includegraphics[width=1in,height=1.25in,clip,keepaspectratio]{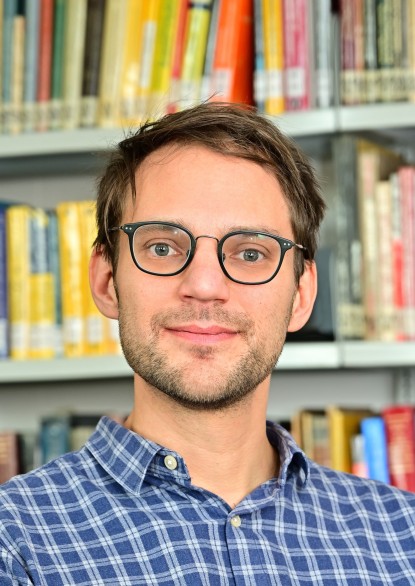}}]{Sebastian Sch\"{o}ps} received the M.Sc. degree in business mathematics and a joint doctoral degree from the Bergische Universität Wuppertal and the Katholieke Universiteit Leuven in mathematics and physics, respectively. He was appointed a Professor of computational electromagnetics with the Technische Universität Darmstadt within the interdisciplinary centre of computational engineering in 2012. Current research interests include coupled multiphysical problems, bridging computer aided design and simulation, parallel algorithms for high performance computing, digital twins, uncertainty quantification and machine learning.
    \end{IEEEbiography}

\end{document}